\documentclass[10pt]{article}

\PassOptionsToPackage{numbers, compress}{natbib}
\usepackage{natbib}
\usepackage[utf8]{inputenc}
\usepackage[a4paper, total={6in, 10in}]{geometry}
\usepackage{graphicx}
\usepackage{indentfirst}
\usepackage{authblk}
\usepackage{lineno}

\usepackage{graphicx}
\usepackage{subcaption}
\usepackage{comment}

\title{Multi-environment lifelong deep reinforcement learning for medical imaging}
\author[2]{Guangyao Zheng}
\author[1]{Shuhao Lai}
\author[2]{Vladimir Braverman}
\author[3]{Michael A. Jacobs}
\author[4]{Vishwa S. Parekh}
\affil[1]{Department of Computer Science, The Johns Hopkins University, Baltimore, MD 21208, USA}
\affil[2]{Department of Computer Science, Rice University, Houston, TX, USA}
\affil[3]{Department Of Diagnostic And Interventional Imaging, McGovern Medical School, UTHealth Houston, Houston, TX, USA}
\affil[4]{Department of Radiology and Radiological Science, Johns Hopkins University School of Medicine, Baltimore, MD 21205, USA}
\affil[5]{University of Maryland Medical Intelligent Imaging (UM2ii) and Department of Diagnostic Radiology and Nuclear Medicine, University of Maryland School of Medicine, Baltimore, MD 21201, USA}

\begin{document}

\maketitle

\begin{abstract} 
Deep reinforcement learning(DRL) is increasingly being explored in medical imaging. However, the environments for medical imaging tasks are constantly evolving in terms of imaging orientations, imaging sequences, and pathologies. To that end, we developed a Lifelong DRL framework, SERIL to continually learn new tasks in changing imaging environments without catastrophic forgetting. SERIL was developed using selective experience replay based lifelong learning technique for the localization of five anatomical landmarks in brain MRI on a sequence of twenty-four different imaging environments. The performance of SERIL, when compared to two baseline setups: MERT(multi-environment-best-case) and SERT(single-environment-worst-case) demonstrated excellent performance with an average distance of $9.90\pm7.35$ pixels from the desired landmark across all 120 tasks, compared to $10.29\pm9.07$ for MERT and $36.37\pm22.41$ for SERT($p<0.05$), demonstrating the excellent potential for continuously learning multiple tasks across dynamically changing imaging environments.
\end{abstract}

\section{Main}
The field of radiology is rapidly adapting artificial intelligence and machine learning techniques for clinical decision support. Deep reinforcement learning (DRL) is a particularly interesting subarea of machine learning methods, as they learn by experiencing and exploring the environment, identical to the human learning process. Diverse fields, including radiology, have benefited significantly from the use of DRL models \cite{mnih2013playing, li2016deep, silver2017mastering, sallab2017deep}. The unique ability of DRL to learn from exploration makes it capable of being used in the Decision Support Systems in radiology. DRL can train to identify and map pathological, anatomical, and structural relationships between distinct radiological images. Researchers are exploring new methods for using DRL in anatomical landmark localization, image segmentation, registration, treatment planning, and assessment in radiology, holding promise for improving radiological diagnosis and treatment \cite{ghesu2017multi,tseng2017deep,maicas2017deep,ma2017multimodal,alansary2018automatic,ali2018lung,alansary2019evaluating,vlontzos2019multiple,Allioui2022MultAgent,Zhang2021wholeprocess,Stember2022drl}.

Reinforcement learning consists of one or more agents exploring and learning about their environments through a finite or infinite set of actions interacting with the environment. The goal of their exploration is to maximize certain reward functions in which the environments provide feedback after agents' every move \cite{sutton2018reinforcement}. To achieve this goal, DRL builds a policy model, which informs an agent of the action to take at a certain state.  DRL leverages the strong learning capabilities of deep learning algorithms, for example, Convolutional Neural Networks (CNN) to learn a representation of the complex environment and simulate the best policy. In the context of medical imaging, the environment will be the 2D images or 3D volumes. However, since medical images are diverse in modality (PET, MRI, X-ray), pathology (benign or malignant tumors), or imaging orientation (axial, coronal, sagittal), the same task may be encountered in very different environments, in terms of difficulty, location, and environment structure. Training multiple DRLs across different anatomical regions and radiological applications would not only increase the space and time complexity of the application but would also be difficult to translate into a clinical workflow as this would result in hundreds of models due to the diversity in body regions and diseases. 

\begin{figure}[htb!]
    \centering
    \includegraphics[width=12cm]{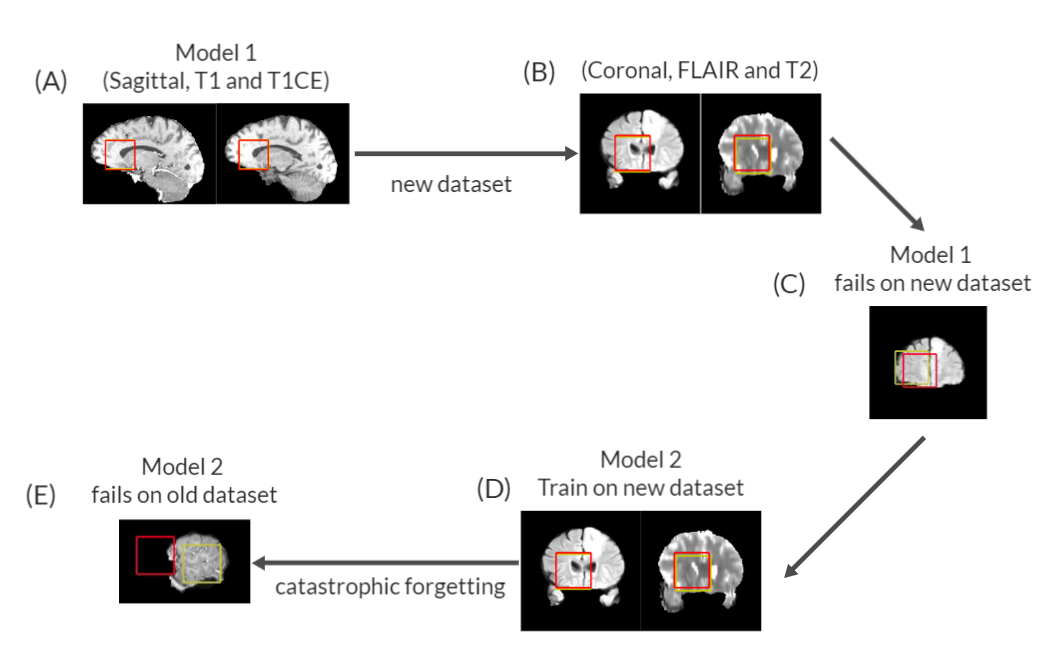}
    \caption{Illustration of catastrophic forgetting in dynamically evolving medical imaging environments. (A) Baseline deep reinforcement learning model trained for ventricle localization in the brain on an environment consisting of T1-weighted pre- and post-contrast enhanced images in Sagittal orientation. (B) The trained model encounters a new environment or new dataset consisting of T2-weighted and FLAIR MRI in the Coronal orientation. (C) The baseline model fails in the new environment due to lack of similar data during training. (D) The baseline model was fine-tuned on the new dataset. (E) The fine-tuning results in catastrophic forgetting where the fine-tuned model no longer works in the original environment.}
    \label{fig:figure1}
\end{figure}

We hypothesize that a single DRL model trained on a diverse set of environments would have an equivalent performance to multiple single environment models trained on each individual environment, therefore resulting in a practical and computationally efficient solution. However, the field of medical imaging is constantly evolving, wherein a new modality or pathology might present itself at a future time point. Therefore, a DRL model that is trained on a predefined set of tasks and environments may not work well on newer unseen tasks and environments. The model can be fine-tuned to work in the newer environment, but that would potentially result in catastrophic forgetting \cite{french1999catastrophic}, meaning the model fails in the original environment it was trained in, as illustrated in Figure \ref{fig:figure1}. Lifelong learning is models that are able to overcome catastrophic forgetting and continuously learn new tasks. Therefore, it is important to integrate lifelong learning capabilities into the existing DRL framework for medical imaging to continually learn different tasks in newer imaging environments without forgetting the old environments. To that end, we developed a selective experience replay based lifelong reinforcement learning framework (SERIL) to train a single model in a system of continuously evolving multiple tasks and multiple environments. Specifically, the SERIL framework uses selective experience replay, introduced by Isele et al. \cite{isele2018selective}, to use the important experience replay buffers, which allows agents to perform lifelong learning.

We trained and evaluated SERIL for the task of anatomical localization of five distinct landmarks in the brain across twenty-four different imaging environments from the 2017 BRATS dataset, consisting of a combination of different MRI sequences, diagnostic pathologies, and imaging orientations. The performance of the SERIL model was compared to two baseline setups: multi-environment (MERT) and single environment (SERT). The MERT setup represents the all-knowing best case model that has access to the complete set of all twenty-four enviornments and five landmarks. In contrast, the SERT setup corresponds to the collection of multiple SERT models, each optimized on a single environment.

\section{Results}

\begin{figure}[htp]
    \centering
    \includegraphics[width=15cm]{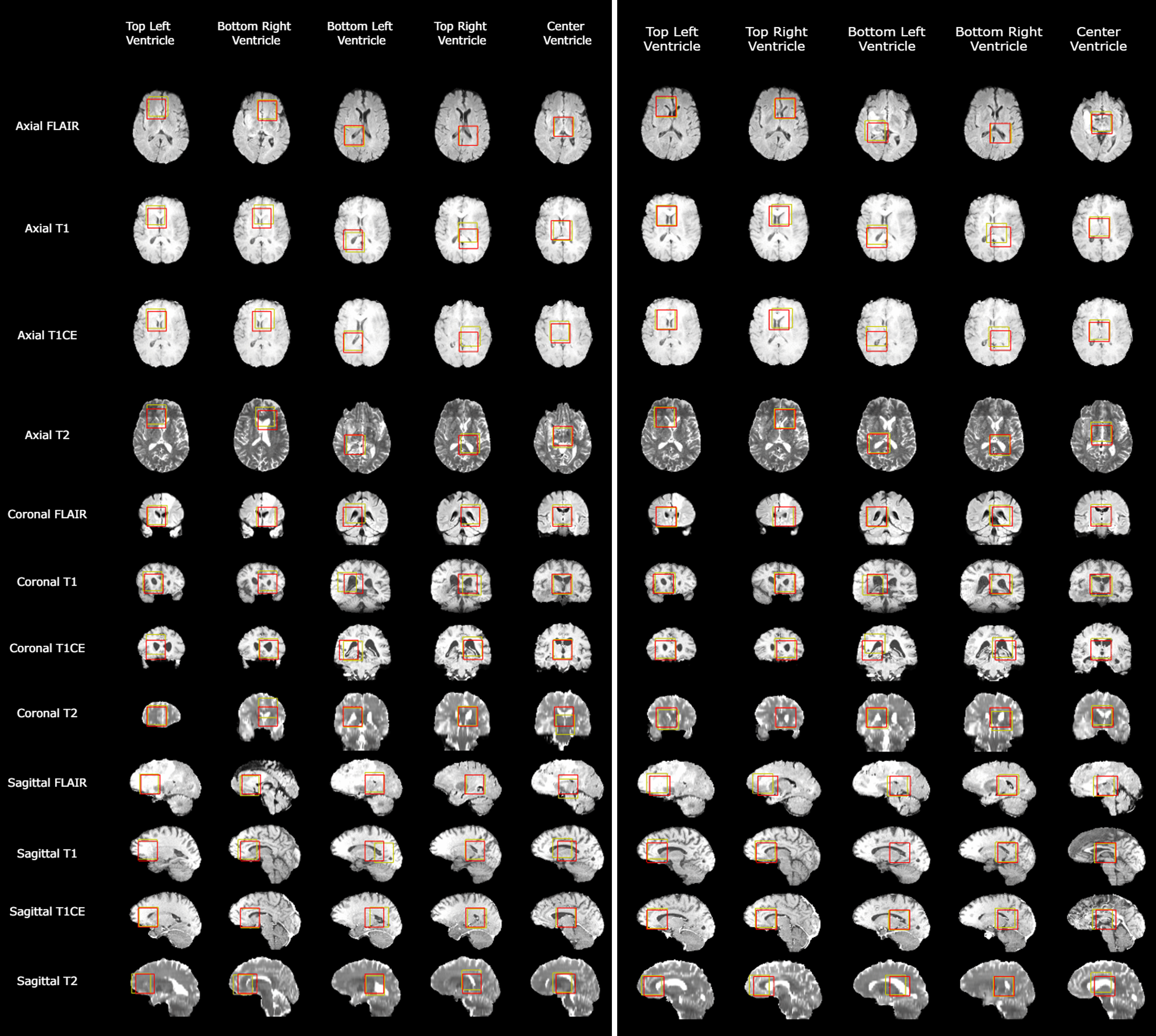}
    \caption{Illustration of the 120 task-environment pairs in our dataset, annotated by the red bounding box is the true landmark location. And the predicted landmark location of the MERT (left) and SERIL (right) models on each of these environment pairs are annotated by the yellow bounding box.}
    \label{fig:figure3}
\end{figure}

We trained SERIL across twenty-four distinct environments and five distinct landmarks (top left ventricle, top right ventricle, bottom left ventricle, bottom right ventricle, and center ventricle). For comparison, we trained twenty-four single-environment multi-agent models (SERT), one for each environment and a single multi-environment (MERT) model. The MERT, SERT, and SERIL models were compared for their performance and generalizability across different environments. Figure \ref{fig:figure3} illustrates the performance of the MERT and SERIL models across 12 different environments (four different imaging sequences: T1, T2, FLAIR, and T1CE and three different imaging orientations: axial, coronal, and sagittal). 

\begin{figure}[htp]
    \centering
    \includegraphics[width=15cm]{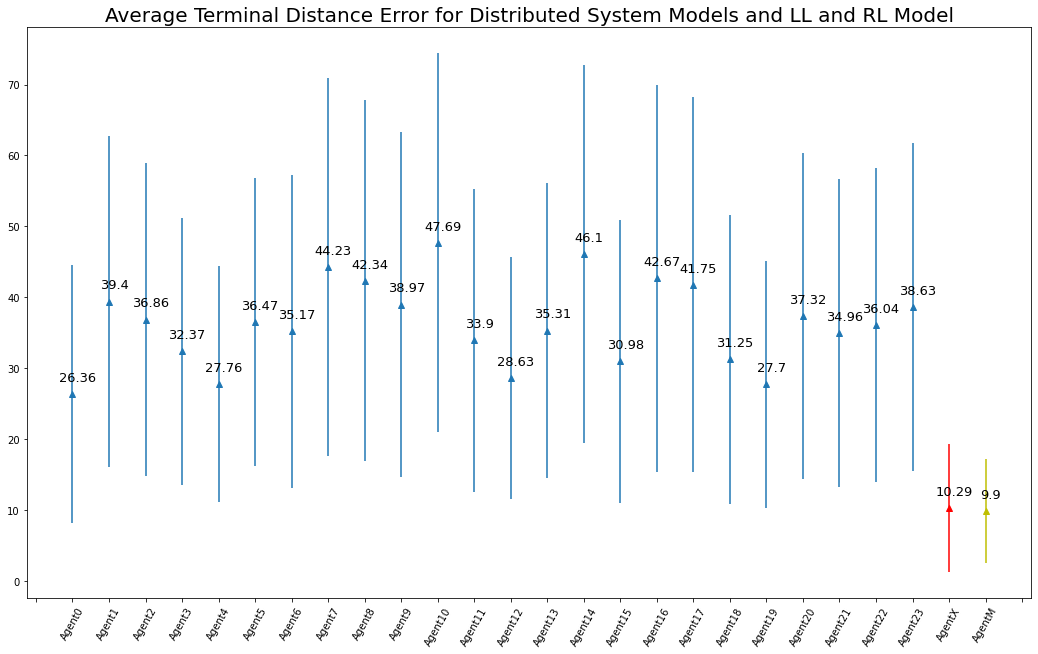}
    \caption{Average Euclidean distance between the predictions of SERT models(Agent $0\sim23$) and MERT(Agent X in red) and SERIL(Agent M in yellow) compared to target landmark locations. The MERT and SERIL outperform all single-environment models in terms of the average distance from the target landmark.}
    \label{fig:figure2}
\end{figure}

The SERIL model demonstrated excellent generalization performance with an average distance of $9.90\pm7.35$ pixels from the desired landmark across all 120 tasks, compared to $10.29\pm 9.07$ for the MERT model. In contrast, the SERT models demonstrated poor generalizability across all 24 environments (120 task-environment pairs) with an average Euclidean distance of $36.37\pm  22.41$ pixels from the desired landmark. 
The comparison of the overall performance of the MERT and SERIL models to each of the twenty-four SERT models is illustrated in Figure \ref{fig:figure2}.

\begin{figure}[htp!]
    \centering
    \includegraphics[width=10cm]{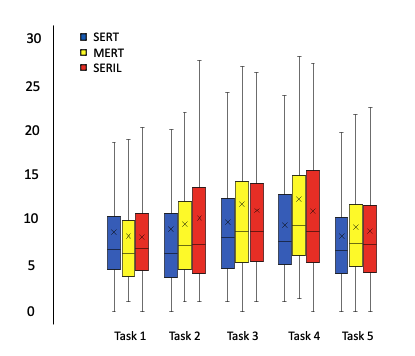}
    \caption{Average Euclidean distance between the predictions of SERT models(Agent $0\sim23$ in blue), MERT (in yellow), and SERIL (in red), compared to target landmark locations. There is no significant difference between the three groups, except task4, where the 24 SERT models outperform the single MERT and SERIL models in terms of the average distance from the target landmark.}
    \label{fig:figure5}
\end{figure}

We compared the SERIL model to the best-performing SERT model for each environment across all task-environment pairs. Figure \ref{fig:figure5} illustrates the overall performance comparison between the twenty-four best-performing SERT models against a single MERT model and a single SERIL model for each landmark localization task, and we see that there is no significant performance difference (except one task, meaning that SERIL is able to learn each task without forgetting previous tasks. These results have been detailed in Table \ref{fig:figure001}.

\begin{table}[]
    \centering
    \caption{Comparative evaluation of SERT, MERT, and SERIL's performances using average Euclidean distance error under five tasks localizing the top left ventricle, top right ventricle, bottom left ventricle, bottom right ventricle, and center ventricle. Pairwise t-test was conducted at the bottom comparing the statistical different between SERT, MERT, and SERIL.}
    \includegraphics[width=15cm]{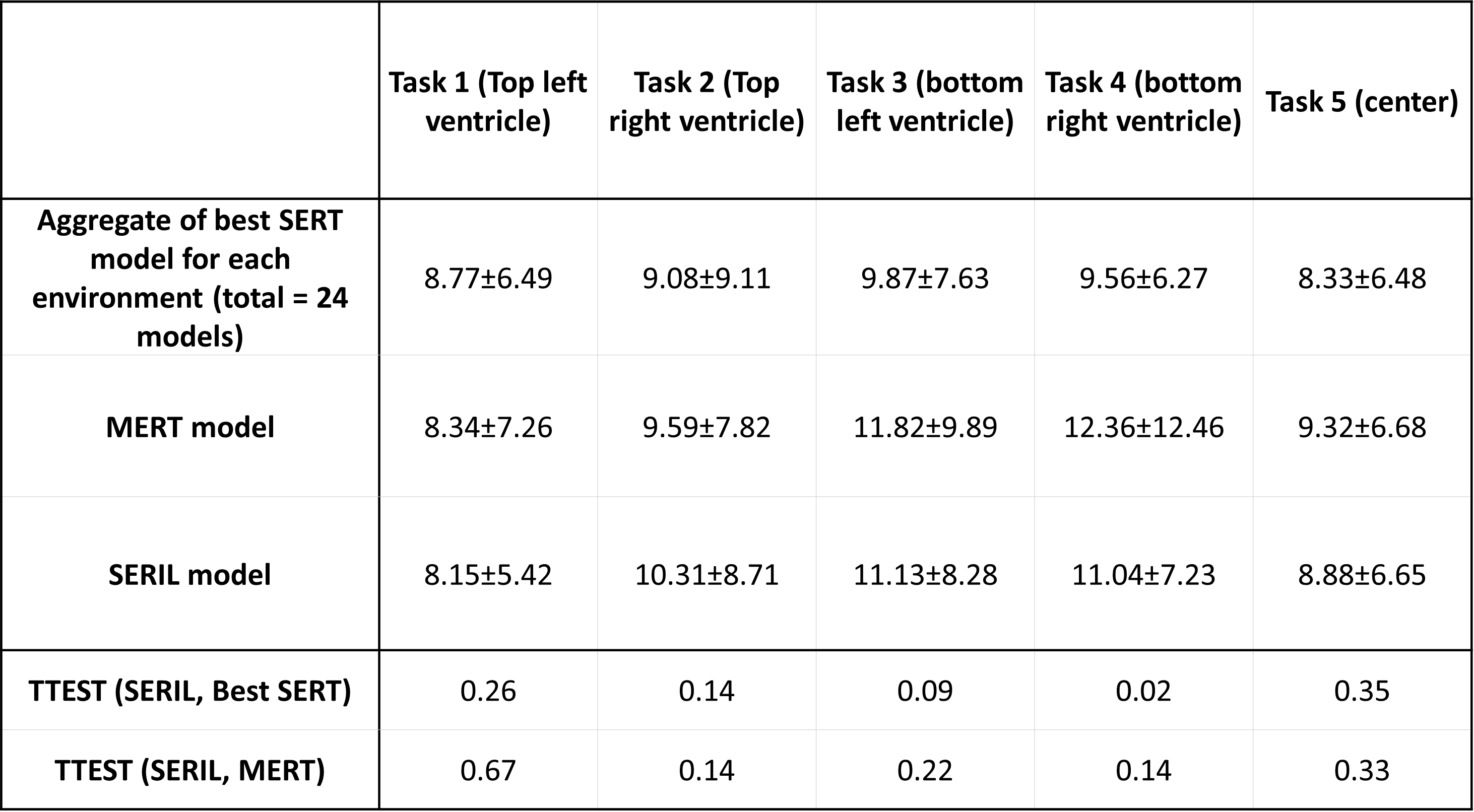}
    
    \label{fig:figure001}
\end{table}

\begin{comment}
We compared the MERT model to the best-performing SERT model for each environment across all task-environment pairs. Figure 6 illustrates the overall performance comparison between the twenty-four best-performing SERT models against a single MERT model for each landmark localization task. Table 1 details the pairwise t-test between the best-performing SERT model and the corresponding MERT model for each task-environment pair, resulting in 120 pairwise comparisons.     
\end{comment}
\begin{comment}
  As shown in Table 1, 107 out of 120 task-environment pairs had no significant difference between the two models. Of the remaining 13 task-environment pairs, the MERT model outperformed the best-performing SERT model in 3 pairs whereas the SERT model outperformed the MERT model in 10 pairs. Overall, the MERT had a comparable or better performance compared to the corresponding best-performing SERT model in $92\%$ of the task-environment pairs.   
\end{comment}

\section{Discussion}
Landmark localization using deep reinforcement learning in radiology has been primarily explored in the context of single image sequences/single modalities without integration of lifelong learning capabilities. Previous works such as DeepNavNet\cite{Edwards2021dn} or Deep Learning-Based Regression and Classification\cite{Noothout2020dr} have shown great results in single environment and single task landmark localization. Similarly, End-to-End Coordinate Regression Model\cite{10.1007/978-3-030-59861-7_63} has also shown great results in localizing anatomical landmarks in a 3D medical imaging setup. Although these frameworks can provide excellent performance given the tasks, they are limited to only the specific image dataset that they trained and tested on. For example, Deep Learning-Based Regression and Classification\cite{Noothout2020dr} trained eight different models to localize eight landmarks for eight different tasks. In \cite{parekh2020multitask}, the authors trained a single model for multiple tasks across different imaging environments, but did not have the lifelong learning capability of dynamically integrating newer environments with time. 

The selective experience replay based multi-task/multi-agent deep reinforcement learning model termed SERIL developed in the work has shown outstanding performance in generalizing across various different image environments. Moreover, they the performance of SERIL was equivalent to best-performing optimized single-environment models(SERT) that are trained specifically for each image environment. 

These frameworks would be very beneficial for translation of medical imaging AI systems, with important practical implications. For example, multiple tasks with multiple environments may need to be trained in a clinical setup. Normally, the models would need to be trained one by one, and the correct model would then need to be selected for computation when an image and a task are identified. Furthermore, the data in a clinical setup may not be aggregated at once. Usually, a patient comes in and their images are acquired. Rather than having to wait for all images to be collected and then train, or retraining the entire model every time a new patient's data is entered, the SERIL framework allows the model to update every time new images are acquired, making use of the data faster and without the huge computational repetition that retraining requires.
 
There are certain limitations to the SERIL framework. For example, they require higher computational complexity to learn multiple tasks in multiple environments at the same time. In the future, we plan to optimize the hyperparameters and the deep neural network to achieve state-of-the-art performance with fewer epochs and iterations and lower the hardware GPU requirements. In conclusion, the SERIL framework demonstrated excellent potential for continuously learning multiple tasks across dynamically changing imaging environments.

\section{Materials and Methods}

\subsection{Deep reinforcement learning}
\begin{figure}[]
    \centering
        \begin{subfigure}{6in}
            \centering
            \subcaption{}
            \includegraphics[width=\linewidth]{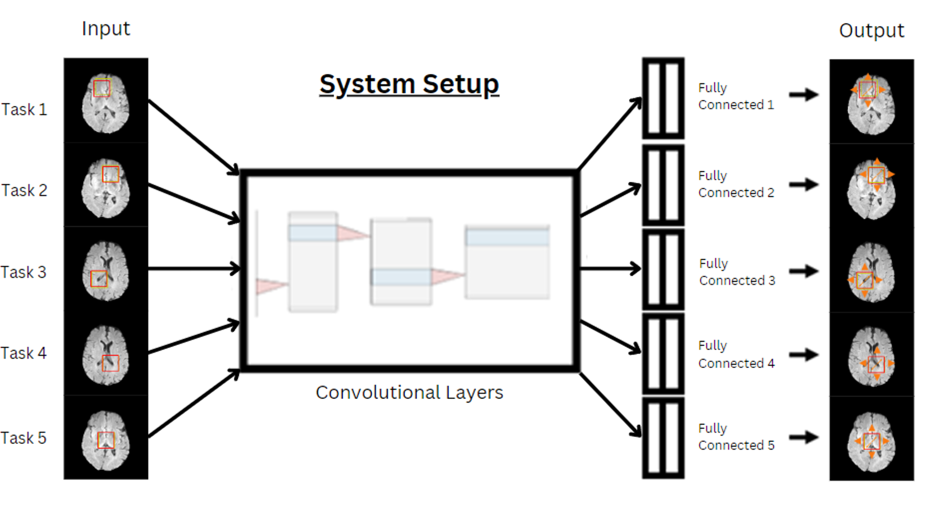}
        \end{subfigure}
     \centering
        \begin{subfigure}{6in}
            \centering
            \subcaption{}
            \includegraphics[width=\linewidth]{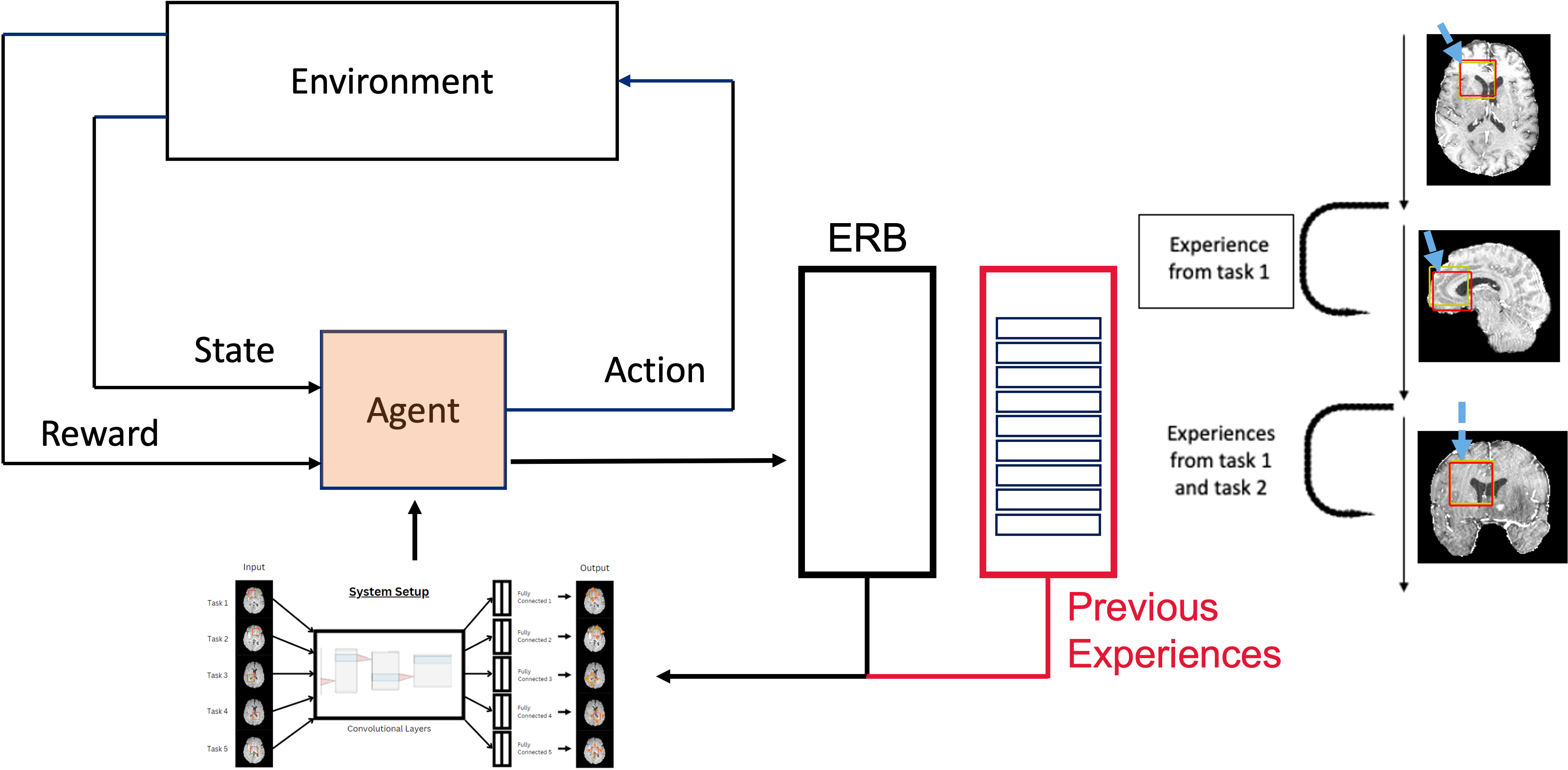}
        \end{subfigure}
    \caption{Illustration of the multi-agent deep reinforcement learning framework. (a) The deep Q-network (DQN) architecture  (b) A schematic of the lifelong deep reinforcement learning setup for training multi-agent deep reinforcement learning models. 
    ERB=Experience Replay Buffer}
\label{fig:figure6}
\end{figure}

In this study, we utilized a deep Q-network (DQN) algorithm to create a multi-agent deep learning framework, which is depicted in Figure \ref{fig:figure6}. The multi-agent DQN model used in this study was modified from previously published works such as \cite{mnih2013playing,alansary2018automatic,vlontzos2019multiple,parekh2020multitask}. The DQN is composed of a central convolutional block with four 3D convolutional layers followed by N fully connected blocks, each comprising 3 layers, as shown in Figure \ref{fig:figure6} (A). The fully connected blocks represent the task-specific blocks for localizing different landmarks. The number of fully connected blocks would dynamically increase with the number of landmarks in the framework. 
The DRL setup for the multi-agent DRL used in this work is shown in Figure \ref{fig:figure6} (B).  There are three crucial components for DRL: the state, the action, and the reward. In the DRL environment, which is the medical images, the agent is represented as a 3D bounding box in the environment. The state is a snapshot of the current location or a sequence of locations of the agent. The actions that allow the agent to interact with the environment are moving in the positive or negative direction along the x, y, and z-axis. Each action will result in the transition from one state to another state. The reward is measured by the difference in distance to the target location before and after a transaction. The DRL agent interacts with the environment according to an $\epsilon$-greedy policy, where an action is taken uniformly at random with probability $\epsilon$ at each step. Otherwise, the action with the highest reward is chosen. An experience replay buffer (ERB) is produced at the end of each training session, containing a collection of state-reward-action-resulting state tuples, denoted as $[s, a, r, s']$, which are generated during the training session using its interaction with the environment across multiple episodes. The DQN algorithm is not only capable of training on the medical images, but also the ERBs as well.

\subsubsection{Multi-environment multi-agent deep reinforcement learning model (MERT)}
A DRL agent's environment corresponds to the 3D imaging volume that the agent operates in and is characterized by the patient's pathology and image acquisition parameters. As a result, there could be potentially many imaging environments that the agent may encounter during deployment, as shown in Figure 5. Therefore, we integrated different imaging environments available during training to train a single multi-environment multi-agent deep reinforcement learning model (MERT). However, the multiplicity of the large set of training environments may potentially result in sub-optimal performance for the model across a certain subset of environments.

\subsubsection{Selective experience replay based multi-task/multi-agent deep reinforcement learning model (SERIL)}
To perform lifelong learning, we implemented a selective experience replay buffer to collect a trajectory of experience samples across the SERIL model's training history. The SERIL model attempts to learn a generalized representation of its current and previous tasks by sampling a batch of experience from both its current task's experience replay buffer (ERB) as well as from its history of previous tasks' experience replays during training. To compare the performance, we trained single-environment multi-agent deep reinforcement learning models (SERT) on each of the environments.

\subsection{Experimental Setup}

\subsubsection{Clinical data}
To assess the performance of the MERT and SERIL and SERT models, we utilized the brain tumor segmentation (BRATS) dataset which includes MRI images in the Axial orientation of 285 patients with various imaging sequences: longitudinal relaxation time (T1) pre-contrast, T1 post-contrast, transverse relaxation time (T2), and Fluid Attenuated Inversion Recovery (FLAIR). We randomly sampled 100 patients out of the 285 total patients for this experiment. Patients also have different pathologies. Out of the 100 patients, 60 patients have high-grade glioma (HGG) and 40 patients have low-grade glioma (LGG). We split the dataset 80:20, resulting in $48 \textrm{ HGG} + 32\textrm{ LGG}$ patients for training and $12 \textrm{ HGG} + 8\textrm{ LGG}$ patients for testing. We also artificially generated the images in the Coronal and Sagittal orientations from the original Axial orientation. Overall, we are able to compile a dataset that consists of twenty-four unique images environments: $4 \textrm{ sequences} \times 2 \textrm{ pathologies} \times 3 \textrm{ orientations}=24$. We used five landmarks (top left ventricle, top right ventricle, bottom left ventricle, bottom right ventricle, and center ventricle) as localization tasks. As a result, we have 120 different task-environment pairs, as illustrated in Figure \ref{fig:figure3}.

\subsubsection{Training protocol}

The MERT model was trained for the localization of all five anatomical landmarks across all twenty-four imaging environments. The MERT model was trained for twenty epochs with a batch size of forty-eight, determined empirically. The agent's state was represented as a bounding box of size 45x45x11 with a frame history length of four. The SERT models were trained for the localization of all five landmarks in each imaging environment, resulting in a total of 24 SERT models. The SERT models were trained for four epochs with a batch size of forty-eight with the same representation for the agent's state as the MERT model. The SERIL model was iteratively trained for the localization of all five anatomical landmarks in one imaging environment at a time. Each iteration was trained for four epochs with a batch size of forty-eight with the same representation for the agent's state as the MERT model. 

All the models were trained and evaluated on NVIDIA DGX-1.

\subsubsection{Performance Evaluation}

The performance metric was set as the terminal Euclidean distance between the agent's prediction and the target landmark. A prediction is better if it is closer to the target landmark in terms of Euclidean distance. Thus we determined empirically that the prediction for a task is adequate if the average Euclidean distance of the prediction is less than 15 pixels away from the target landmark. We also performed pairwise t-tests to compare the performance of the MERT model with the SERT models. The p-value for statistical significance was set to $p \le 0.05$.

\bibliographystyle{unsrtnat}
\bibliography{parekh22_MERT}

\end{document}